\documentclass{article}
\usepackage{ijcai26}

\usepackage{times}
\usepackage{soul}
\usepackage{url}
\usepackage[hidelinks]{hyperref}
\usepackage[utf8]{inputenc}
\usepackage[small]{caption}
\usepackage{graphicx}
\usepackage{amsmath}
\usepackage{amsthm}
\usepackage{amssymb}
\usepackage{booktabs}
\usepackage{algorithm}
\usepackage{algorithmic}
\usepackage[switch]{lineno}
\usepackage{xcolor}

\title{AwakeForest: An Interactive Geospatial Platform for Large-Scale Forest Imagery\thanks{To appear at IJCAI-ECAI 2026 (Demo Track).}}

\author{
Suraj Prasai$^1$
\and
Kangning Cui$^1$
\and
Rongkun Zhu$^2$
\and
Sarra Alqahtani$^1$
\and
Ying Zhang$^1$
\and \\
Victor Paul Pauca$^1$
\and
Miles R. Silman$^{1}$
\And
Fan Yang$^1$\\
\affiliations
$^1$Wake Forest University\\
$^2$Hong Kong Baptist University\\
\emails
\{prass25, cuij, alqahtas, zhangyi, paucavp, silmanmr, yangfan\}@wfu.edu,
csrkzhu@comp.hkbu.edu.hk
}

\begin{document}

\maketitle

\begin{abstract}

Forest imagery analysis often involves multiple tightly coupled vision tasks, which must be performed under substantial variation in geographic regions, sensors, and acquisition conditions. However, practitioners often lack a \textit{unified} tool that is geospatial-native, cloud-optimized, and ML-integrated for end-to-end workflows spanning annotation, prediction, visualization, and downstream analysis at scale. We present \texttt{AwakeForest}, an interactive end-to-end platform designed for large-scale forest imagery that integrates model-assisted inference, automatic annotation, and human-in-the-loop refinement within a single workflow. Our platform supports \textit{plug-and-play} integration of pretrained models and enables scalable interaction with forest imagery ranging from standard aerial scenes to large orthomosaics that can span several gigabytes to hundreds of gigabytes. \texttt{AwakeForest} produces analysis-ready outputs that can be directly used for downstream analysis and to support iterative model and annotation updates on new scenes. We demonstrate the system on the PALMS dataset and illustrate how \texttt{AwakeForest} supports an end-to-end workflow for practical forest management and analysis.

\end{abstract}

\section{Introduction}


Forest imagery analysis commonly requires multiple vision tasks, including object detection, spatial delineation, and quantitative analysis, for ecological monitoring and forest management~\cite{camalan2022change,de2025remote,cui2025efficient}. In practice, these tasks are tightly coupled and need to be conducted within a single workflow. However, constructing high-quality labeled datasets is costly, and domain shifts across regions, sensors, and acquisition conditions further limit the reuse of existing annotations and models. As a result, practitioners often need to repeatedly adapt models and rebuild datasets with limited integrated tooling.

\begin{figure}[t]
  \centering
  \includegraphics[
    height=0.30\textheight,
    keepaspectratio
  ]{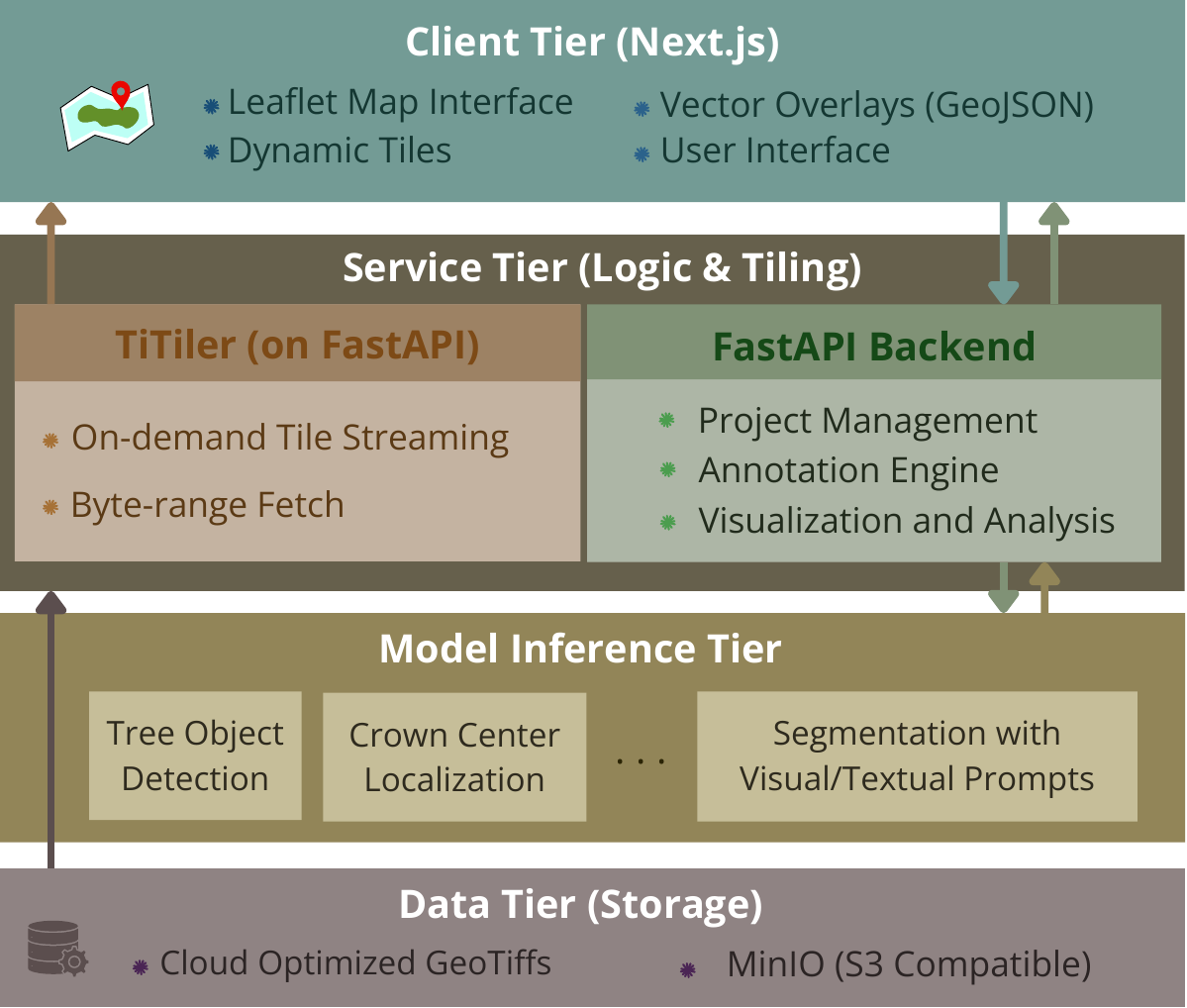}
  \caption{Overview of \texttt{AwakeForest} design.}
  \label{fig:system}
\end{figure}

\begin{figure*}[t]
  \centering
  \includegraphics[
    width=0.99\textwidth,
    height=0.27\textheight,
    keepaspectratio
  ]{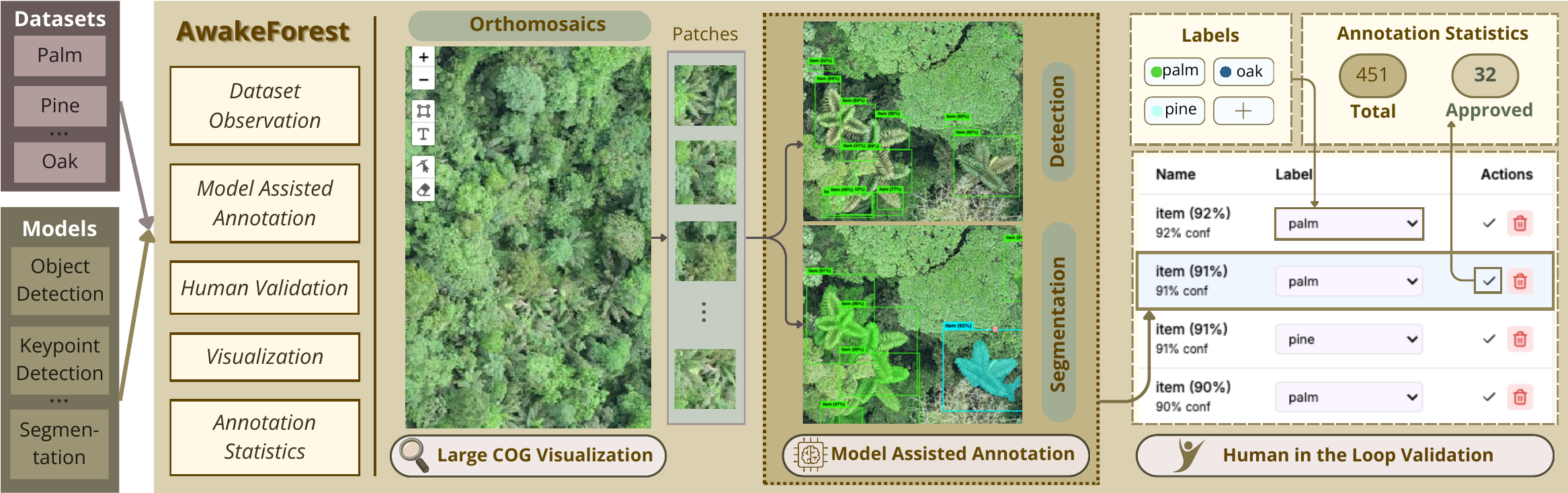}
  \caption{
  The \texttt{AwakeForest} workflow and interface. \textit{Left}: Project configuration panel where users link dataset and select registered models. \textit{Right}: Annotation workflow showing: large COG orthomosaic in global view, patch-based subdivision, model-assisted annotation using YOLO and YOLO+SAM2, human-in-the-loop label assignment and verification, and annotation statistics.
  }
  \label{fig:workflow}
\end{figure*}

Several existing platforms support model-assisted labeling and integration of pretrained models. Tools such as CVAT~\cite{cvat}, AnyLabeling~\cite{anylabeling}, and Label Studio~\cite{LabelStudio} allow users to load detection or segmentation models to assist manual annotation and are effective for generic object annotation workflows and conventional computer vision datasets. However, they are not specifically designed for forest and ecological analysis, where imagery is often high-resolution and spatially extensive. In many scenarios, ecologists work with large orthomosaic images produced by automated reconstruction pipelines, where a single image can range from several to hundreds of gigabytes and objects of interest may be small and densely distributed. In these settings, annotation is tightly coupled with analysis, requiring iterative integration of detection, segmentation, counting, and domain-driven refinement rather than treating labeling as a standalone preprocessing step.


\texttt{AwakeForest} addresses this gap by focusing on large-scale imagery and practical forest management. We provide a unified interactive workflow (Figure~\ref{fig:system}) that pairs automated predictions with human-in-the-loop refinement within a consistent geospatial context. By tightly integrating geospatial data, model inference, annotation, and visualization, our end-to-end pipeline enables users to move seamlessly from large imagery exploration to verified, analysis-ready outputs.

\section{System Overview}
\label{sec:overview}

\subsection{High-Level System Architecture}
\label{sec:overview-architecture}

\texttt{AwakeForest} adopts a modular, service-oriented architecture that separates user interaction, workflow coordination, model inference, and data management (see Figure~\ref{fig:system}). This separation allows system components to evolve independently while maintaining a coherent end-to-end workflow.

The \textit{client tier} provides a user interface for managing datasets, models, and projects. Users can register datasets by specifying object storage endpoints and bucket details, add detection or segmentation models hosted locally or remotely by providing API endpoints, and create projects that link datasets with models. The \textit{service tier} serves as a central coordinator responsible for user authentication, project organization, annotation management, and orchestration of inference requests across tasks. The \textit{model inference tier} consists of independent services registered by users through model endpoints and task type specifications. The \textit{data tier} serves large datasets through scalable object storage with on-demand tile streaming, separating raw imagery from spatial metadata and annotations to support interactive analysis.

\subsection{Integrated Geospatial Annotation Workflow}
\label{sec:overview-workflow}

\texttt{AwakeForest} organizes forest imagery analysis through a project-based structure that links geospatial datasets with external model services for interactive annotation, as shown in Figure~\ref{fig:workflow}. Each project connects a dataset with the registered models and provides two complementary views of the data.

The \textit{global view} displays the entire orthomosaic, allowing users to navigate with pan and zoom controls while viewing annotation overlays for spatial context and progress tracking across large forest scenes in real time. The \textit{patch-level annotation view} divides the dataset into fixed-size image patches where users can select models from a drop-down menu to generate predictions on the current patch; manually select and annotate objects missed by the models; assign labels and verify annotations for quality control and consistency; and navigate between patches to systematically annotate the dataset. 

To ensure consistency across automated predictions and user edits, \texttt{AwakeForest} maintains a unified internal representation of intermediate results based on geospatial coordinates. By decoupling annotations from specific image tiles and patch boundaries, objects of interest are stored with absolute spatial attributes and standardized geometry formats. This design supports stable indexing, cross-model comparison, and incremental updates during refinement. It allows annotations to be managed and overlaid independently across diverse datasets or imagery sources with different resolutions, enabling outputs from various models to be combined, inspected, and refined within a cohesive spatial context. 

\subsection{Implementation Details}
\label{sec:overview-implementation}

The platform is implemented using Next.js~\cite{nextjs14} for the \textit{client} tier and FastAPI~\cite{fastapi} for the \textit{service} tier, with Leaflet~\cite{leaflet} providing interactive geospatial map visualization. Data persistence and operations are managed through Supabase~\cite{supabase}, which provides spatial indexing via PostGIS (a PostgreSQL extension for geospatial queries) ~\cite{postgis} and JWT-based authentication (JSON Web Tokens for secure, stateless session management).

Cloud Optimized GeoTIFFs (COGs)~\cite{cog} are stored in MinIO~\cite{minio} object storage and served through dynamic tiling via TiTiler~\cite{titiler} integration. The service tier handles on-demand tile generation based on client viewport requests, enabling efficient streaming of image data at the scale required for interactive exploration of large datasets without preprocessing.

Model inference services are registered as external FastAPI endpoints defined by task type, such as detection or segmentation. This approach allows the platform to invoke remote inference during patch-level annotation without coupling inference logic to the core application. The architecture supports integration of various vision models including YOLO ~\cite{yolo,ultralytics} detectors and SAM~\cite{sam2,sam3} models for object detection, segmentation, and automatic labeling tasks.

\subsection{Comparison with Existing Tools}
\label{sec:comparison}

Existing annotation tools such as Label Studio, AnyLabeling, and CVAT support generic image-centric labeling with optional AI-assisted proposals, but they do not treat geospatial coordinates and large orthomosaics as first-class primitives. Conversely, geographic information system (GIS) platforms such as QGIS~\cite{QGIS_software} are geospatial-native environments designed for spatial data management, indexing, and map-based analysis, yet they lack an integrated interactive ML annotation workflow with plug-and-play model inference capabilities. \texttt{AwakeForest} combines geospatial-first data management, cloud-optimized orthomosaic access, and model-integrated interactive labeling within a unified system architecture and workflow (Table~\ref{tab:comparison}).

\begin{table}[t] 
    \centering 
    \footnotesize 
    \addtolength{\tabcolsep}{-3.5pt} 
    \begin{tabular}{l c c c c} 
    \toprule 
    \textbf{Tool} & \textbf{\shortstack{Geospatial\\Native}} & \textbf{\shortstack{Spatial\\Indexing}} & \textbf{\shortstack{Cloud \\Optimized}} & \textbf{\shortstack{AI-Assisted \\Labeling}} \\ 
    \midrule 
    Label Studio & $\times$ & $\times$ & $\times$ & \checkmark \\ 
    CVAT         & $\times$ & $\times$ & $\times$ & \checkmark \\ 
    AnyLabeling  & $\times$ & $\times$ & $\times$ & $\checkmark$ 
    \\
    QGIS         & \checkmark & \checkmark & \checkmark & $\times$ \\ 
    \textbf{AwakeForest}  & \textbf{\checkmark} & \textbf{\checkmark} & \textbf{\checkmark} & \textbf{\checkmark} \\ 
    \bottomrule 
    \end{tabular} 
    \caption{Comparison with popular annotation tools.}
    \label{tab:comparison}
\end{table}

\section{Interactive Workflow and Demonstration}
\label{sec:workflow}

We show the capabilities of \texttt{AwakeForest} on the PALMS dataset~\cite{cui2025detection}, which provides a realistic case for interactive forest imagery analysis at scale. PALMS consists of high-resolution UAV-derived orthomosaics covering large, spatially continuous forest regions with dense and heterogeneous canopy structure. These characteristics require joint handling of labeling, object detection and localization, segmentation, and spatial reasoning, which are well suited for illustrating how \texttt{AwakeForest} supports end-to-end interaction from initial inference to refined, analysis-ready results.

Within \texttt{AwakeForest}, PALMS is loaded as large orthomosaics and interactively explored at multiple spatial scales. Users first identify regions of interest corresponding to forest patches or survey areas and initiate object searching using pretrained detection models. These models generate initial proposals for palm instances, including bounding boxes and candidate crown centers. Optionally, users can invoke SAM-based segmentation using these predicted boxes as prompts to obtain high-resolution instance masks for each detected palm. This enables quick conversion from coarse localization to detailed crown delineation without manual mask drawing.

Users can also invoke prompt-based segmentation for automatic labeling using either bounding boxes or text prompts. For PALMS, text prompts such as ``\textit{palm}'' or ``\textit{palm tree}'' can be used to retrieve candidate instances and generate corresponding masks for rapid initialization. With box prompts, the system produces a mask for the specified region and leverages the few-shot capability of SAM3 to suggest additional masks for visually similar crowns beyond the initial box. This interaction supports rapid discovery and correction of missed instances and boundary errors, reducing the need for manual annotation from scratch and improving overall efficiency.

\begin{figure}[t]
  \centering
  \includegraphics[width=\linewidth]{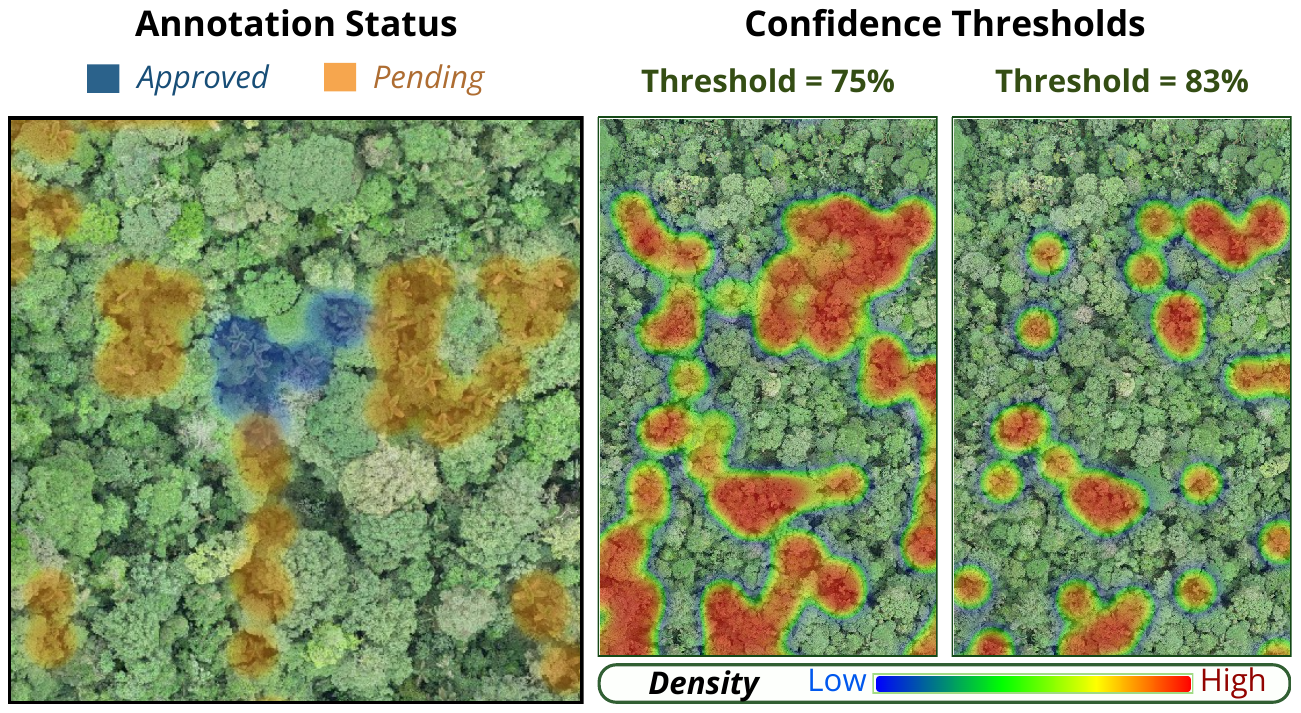}
  \caption{The analysis-ready outputs. Orthomosaic-level views for status verification and confidence filtering; prediction densities are computed via kernel density estimation over instance locations.}
  \label{fig:visualization}
\end{figure}

Throughout the demonstration, automated predictions and user refinements are maintained within a single project context. Verified results can be promoted to finalized annotations, while intermediate predictions remain traceable to their originating models. Figure~\ref{fig:visualization} summarizes the analysis-ready outcomes at the orthomosaic level, combining verification status with confidence-thresholded prediction densities. Users can filter predictions by confidence, and the retained instance locations are summarized via kernel density estimation to reveal spatial clusters and coverage patterns. The finalized outputs are then used to derive spatial summaries such as palm center distributions and instance counts over selected regions~\cite{cui2025efficient,zhu2025orthomosaics}.

\section{Discussion and Conclusion}
\label{sec:discussion}

\texttt{AwakeForest} is an interactive platform that integrates model-assisted inference with human-in-the-loop refinement for large-scale forest imagery analysis. As a workflow-oriented platform, the quality of intermediate and final results depends on the suitability of the employed pretrained models, particularly under substantial domain shift or visually complex forest conditions, where additional user refinement may be required~\cite{tuia2016domainadaptation,quinonero2008datasetshift}. Interactive analysis over very large orthomosaics also entails practical computational considerations~\cite{fails2003interactive,amershi2014power}, such as inference latency and visualization overhead, which are inherent to large-scale geospatial data processing~\cite{gorelick2017earthengine}.

This work focuses on demonstrating the platform on RGB forest imagery in a representative forest management scenario and does not aim to provide quantitative benchmarks on predictive accuracy or annotation efficiency. Extensions to additional sensing modalities, including multispectral imagery~\cite{hansen2013forestcover}, LiDAR~\cite{lefsky2002lidarbiomass}, and temporal data~\cite{banskota2014landsatreview}, as well as collaborative and multi-user workflows~\cite{palomino2017review,ma2022customizable}, are left as future work. Within its current scope, \texttt{AwakeForest} illustrates how an integrated, end-to-end workflow can support detection, segmentation, spatial summarization, and iterative refinement in a unified environment for practical forest management tasks.


\appendix



\section*{Acknowledgments}
We thank the reviewers for their constructive feedback.
We also thank the Center for Amazonian Scientific Innovation (CINCIA) for providing data access. Our code is available at: \url{https://github.com/spygaurad/AwakeForest}, and the demo video can be viewed at: \url{https://youtu.be/Z0EZj1PPDv4}. 
 
\section*{Contribution Statement}
Suraj Prasai and Kangning Cui contributed equally to this work. Fan Yang served as the corresponding author.

\bibliographystyle{named}
\bibliography{ijcai26}

\end{document}